\documentclass{article}
\usepackage{spconf,amsmath,graphicx,hyperref}
\usepackage{hyperref}
\usepackage{url}
\usepackage{algorithmic}
\usepackage{algorithm}
\usepackage{tikz}
\usepackage{subcaption} % For side-by-side diagrams
\usepackage{pgfplots}   % For plotting
\usepackage{booktabs}      % For professional table rules
\usepackage{multirow}      % For merging rows
\usepackage{tabularx}      % For full-width tables
\usepackage[most]{tcolorbox}
\usepackage{markdown}
\usepackage{bm}

\pgfplotsset{compat=1.18} % Ensure compatibility with latest version

\title{Investigating Batch Inference in a Sequential Monte Carlo Framework for Neural Networks} %
\name{Andrew Millard$^{*}$$^{\dagger}$, Joshua Murphy$^{*}$$^{\dagger}$, Peter L Green$^{**}$, Simon Maskell$^{*}$
\thanks{$^\dagger$These authors contributed equally. \\
AM and JM were funded by a Research Studentship jointly funded by the EPSRC Centre for Doctoral Training in Distributed Algorithms EP/S023445/1. SM was funded by Dstl in collaboration with the Royal Academy of Engineering via "Dstl-RAEng Research Chair in Information Fusion" under task RQ0000040616.}}
\address{Department of Electronics and Electrical Engineering, University of Liverpool$^{*}$ \\Department of Mechanical and Aerospace Engineering, University of Liverpool$^{**}$}
\author{\name Andrew Millard\email amill212@liverpool.ac.uk \\
      \addr University of Liverpool \\
      \ANDt
      \name Joshua Murphy \email jmurph98@liverpool.ac.uk \\
      \addr University of Liverpool \\
      \ANDt
      \name Peter L Green \email plgreen@liverpool.ac.uk\\
      \addr University of Liverpool  \\
      \ANDt
      \name Simon Maskell  \email smaskell@liverpool.ac.uk\\
      \addr University of Liverpool \\
    }
\begin{document}
%\ninept
%
\maketitle
\begin{abstract}
Bayesian inference allows us to define a posterior distribution over the weights of a generic neural network (NN). Exact posteriors are usually intractable, in which case approximations can be employed. One such approximation - variational inference - is computationally efficient when using mini-batch stochastic gradient descent as subsets of the data are used for likelihood and gradient evaluations, though the approach relies on the selection of a variational distribution which sufficiently matches the form of the posterior. Particle-based methods such as Markov chain Monte Carlo and Sequential Monte Carlo (SMC) do not assume a parametric family for the posterior by typically require higher computational cost. These sampling methods typically use the full-batch of data for likelihood and gradient evaluations, which contributes to this computational expense. We explore several methods of gradually introducing more mini-batches of data (data annealing) into likelihood and gradient evaluations of an SMC sampler. We find that we can achieve up to $6\times$ faster training with minimal loss in accuracy on benchmark image classification problems using NNs.
\end{abstract}

\begin{keywords}
    Sequential Monte Carlo, Batch Inference, Neural Networks 
\end{keywords}

\section{Introduction}

Bayesian inference offers a principled way to handle uncertainty in Neural Networks (NN) by treating parameters as random variables and inferring, from training data, their posterior distribution. While this posterior is often intractable for NNs, it can be approximated through parametric approaches such as variational inference (VI) \cite{graves2011practical} or non-parametric approaches such as Markov chain Monte Carlo (MCMC)\cite{neal2012bayesian}. VI with stochastic gradient descent can be applied to large datasets as it only evaluates small mini-batches (MBs) of data per optimisation step. However, VI relies on the selection of an appropriate family of variational distribution (e.g. Gaussian) which can be difficult for the complex posteriors that arise when NNs are considered. MCMC produces asymptotically exact samples but is computationally intensive and has been shown to scale poorly to high-dimensional models~\cite{betancourt2015fundamental}.

Sequential Monte Carlo (SMC) samplers \cite{del2006sequential} are a parallelisable particle-based methods which mitigates some of the computational expense of MCMC \cite{rosato2023log}, though can remain prohibitively expensive for large NNs~\cite{bengtsson2008curse}. Part of this computational cost comes from the use of the full-batch of data in likelihood and gradient evaluations \cite{betancourt2015fundamental}. Stochastic gradient (and likelihood) estimates utilise a subset of the data to approximate the gradient that would be obtained from the Full-batch \cite{chen2014stochastic} (where larger batches give better approximations).

%We propose to gradually increase the size of the mini-batch (data annealing) used in these stochastic gradient evaluations which results in faster sampling of NN posteriors using SMC samplers with little degradation in terms of loss and accuracy on example classification tasks. 

We propose to gradually increase the size of the subset of data from which we calculate the stochastic estimates by appending MBs to the current data subset, which is known as \emph{data annealing} (DA). We investigate two forms; one where the data is introduced via a predefined schedule, and one where we use posterior statistics to inform us when a new MB should be annealed in. The paper is structured as follows: Section~\ref{sec:SMC} introduces SMC samplers, Section~\ref{sec:batch} discusses MB inference and annealing schemes with Section~\ref{sec:discussion} presenting experiments on image classification tasks along with the discussion and conclusions.

%We investigate both approaches to estimating the gradient of the target distribution and data annealing, whereby data is introduced into the likelihood until posterior statistics are judged to have converged. 
\section{Sequential Monte Carlo Samplers}\label{sec:SMC}

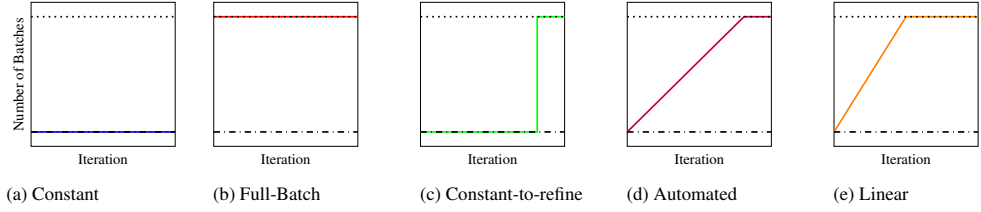
\begin{figure*}[h!]
    \centering
    \resizebox{0.8\textwidth}{!}{
    \begin{minipage}{\textwidth}
    \begin{subfigure}[b]{0.19\textwidth}
        \begin{tikzpicture}
            \begin{axis}[
                xlabel={Iteration},
                ylabel={Number of Batches},
                xmin=0, xmax=10,
                ymin=0, ymax=20,
                xtick=\empty, ytick=\empty,
                width=4cm, height=4cm,
                grid=major,
                label style={font=\scriptsize} % Reduced font size for axis labels
            ]
            \addplot[color=blue, thick] coordinates {(0,2) (2,2) (4,2) (6,2) (8,2) (10,2)};
            \addplot[dotted, thick] coordinates {(0,18) (10,18)};
            \addplot[dashdotted, thick] coordinates {(0,2) (10,2)};
            \end{axis}
        \end{tikzpicture}
        \caption{Constant}
    \end{subfigure}
    \begin{subfigure}[b]{0.19\textwidth}
        \begin{tikzpicture}
            \begin{axis}[
                xlabel={Iteration},
                xmin=0, xmax=10,
                ymin=0, ymax=20,
                xtick=\empty, ytick=\empty,
                width=4cm, height=4cm,
                grid=major,
                label style={font=\scriptsize} % Reduced font size for axis labels
            ]
            \addplot[color=red, thick] coordinates {(0,18) (10,18)};
            \addplot[dotted, thick] coordinates {(0,18) (10,18)};
            \addplot[dashdotted, thick] coordinates {(0,2) (10,2)};
            \end{axis}
        \end{tikzpicture}
        \caption{Full-Batch}
    \end{subfigure}
    \begin{subfigure}[b]{0.19\textwidth}
        \begin{tikzpicture}
            \begin{axis}[
                xlabel={Iteration},
                xmin=0, xmax=10,
                ymin=0, ymax=20,
                xtick=\empty, ytick=\empty,
                width=4cm, height=4cm,
                grid=major,
                label style={font=\scriptsize} % Reduced font size for axis labels
            ]
            \addplot[color=green, thick] coordinates {(0,2) (8,2) (8.1,2) (8.1,18) (10,18)};
            \addplot[dotted, thick] coordinates {(0,18) (10,18)};
            \addplot[dashdotted, thick] coordinates {(0,2) (10,2)};
            \end{axis}
        \end{tikzpicture}
        \caption{Constant-to-refine}
    \end{subfigure}
    \begin{subfigure}[b]{0.19\textwidth}
        \begin{tikzpicture}
            \begin{axis}[
                xlabel={Iteration},
                xmin=0, xmax=10,
                ymin=0, ymax=20,
                xtick=\empty, ytick=\empty,
                width=4cm, height=4cm,
                grid=major,
                label style={font=\scriptsize} % Reduced font size for axis labels
            ]
            \addplot[color=purple, thick] coordinates {(0,2) (8.1,18) (10,18)};
            \addplot[dotted, thick] coordinates {(0,18) (10,18)};
            \addplot[dashdotted, thick] coordinates {(0,2) (10,2)};
            \end{axis}
        \end{tikzpicture}
        \caption{Automated}
    \end{subfigure}
    \begin{subfigure}[b]{0.19\textwidth}
        \begin{tikzpicture}
            \begin{axis}[
                xlabel={Iteration},
                xmin=0, xmax=10,
                ymin=0, ymax=20,
                xtick=\empty, ytick=\empty,
                width=4cm, height=4cm,
                grid=major,
                label style={font=\scriptsize} % Reduced font size for axis labels
            ]
            \addplot[color=orange, thick] coordinates {(0,2) (5,18) (10,18)};
            \addplot[dotted, thick] coordinates {(0,18) (10,18)};
            \addplot[dashdotted, thick] coordinates {(0,2) (10,2)};
            \end{axis}
        \end{tikzpicture}
        \caption{Linear}
    \end{subfigure}
    
    % Legend
    \begin{tikzpicture}
      \begin{axis}[
        hide axis,
        width=\linewidth,
        height=1.2cm,        % increase from 0.5cm
        scale only axis,     % size applies to axis box only
        legend columns=-1,
        legend style={draw=none, column sep=1cm, anchor=north}
      ]
        \addlegendimage{dotted, thick}
        \addlegendentry{Dotted Line: Full Batch}
        \addlegendimage{dashdotted, thick}
        \addlegendentry{Dash-Dotted Line: Single Batch}
      \end{axis}
    \end{tikzpicture}
    \end{minipage}
    }
    \caption{Visualisation of batching schemes.}
    \label{fig1:anneal_schemes}
\end{figure*}

SMC samplers are an importance sampling method which can be applied to Bayesian inference problems to generate samples from posterior distributions. SMC samplers use a set of particles $\{ \bm{\theta}_k^{(j)} \}_{j=1}^J$ with importance weights $\{ \mathbf{w}_k^{(j)} \}_{j=1}^J$ to target, in the current work, a NN posterior $\pi(\bm{\theta})\propto p(\mathbf{y}_{1:N}|\bm{\theta})q_0(\bm{\theta})$ over $K$ iterations, where $p(\mathbf{y}_{1:N}|\bm{\theta})$ is the likelihood of the NN, $N$ is the size of the dataset and $q_0(\bm{\theta})$ is the prior. In a standard SMC implementation, particles are first sampled from the prior and weighted
\begin{equation}
    \bm{\theta}_0^{(j)} \sim q_0(\bm{\theta}), \qquad 
    \mathbf{w}_0^{(j)} = \frac{\pi(\bm{\theta}^{(j)}_0)}{q_0(\bm{\theta}^{(j)}_0)}, 
    \qquad j=1,\dots,J. 
    \label{equ:init_weight}
\end{equation}
At iteration $k$, particles are propagated via a Markov kernel~(MK)
\begin{equation}
    \bm{\theta}_k^{(j)} \sim q_k(\bm{\theta}^{(j)}_{k}|\bm{\theta}^{(j)}_{k-1}), 
    \qquad j=1,\dots,J , 
    \label{equ:proposal}
\end{equation}
and reweighted as
\begin{equation}
    \mathbf{w}^{(j)}_k = \mathbf{w}^{(j)}_{k-1} 
    \frac{\pi(\bm{\theta}^{(j)}_k)}{\pi(\bm{\theta}^{(j)}_{k-1})}
    \frac{L_k(\bm{\theta}^{(j)}_{k-1}|\bm{\theta}^{(j)}_{k})}{q_k(\bm{\theta}^{(j)}_{k}|\bm{\theta}^{(j)}_{k-1})},
    \qquad j=1,\dots,J, 
    \label{equ:weight_update}
\end{equation}
where $L_k$ is known as the L-kernel \cite{del2006sequential}. Weights are normalised to obtain
\begin{equation}
    \Tilde{\mathbf{w}}_k^{(j)} = \frac{\mathbf{w}_k^{(j)}}{\sum_{j=1}^J \mathbf{w}_k^{(j)}},
    \qquad j=1,\dots,J, 
    \label{equ:normalise}
\end{equation}
and resampling \cite{douc2005comparisonresamplingschemesparticle} occurs when the effective sample size 
\begin{equation}
    J_\mathrm{eff} = \frac{1}{\sum_{j=1}^J (\Tilde{\mathbf{w}}_k^{(j)})^2},
    \qquad j=1,\dots,J , 
    \label{equ:jeff}
\end{equation}
drops below $J/2$ where, after resampling, weights are reset to $1/J$. This mitigates particle degeneracy, where a single particle dominates the weights, degrading the posterior estimate. The effects of resampling on degeneracy have been studied previously \cite{douc2005comparisonresamplingschemesparticle, godsill2019particle}. Here, we use multinomial resampling. Estimates of functions on the distribution, $f(\bm{\theta})$, are realised by
\begin{equation}
    \mathbb{E}_\pi[f(\bm{\theta})] \approx \sum_{j=1}^{J} \Tilde{\mathbf{w}}^{(j)}_{k} f(\bm{\theta}^{(j)}_{k}).
\end{equation}

\subsection{Markov and L-Kernels}\label{subsec:proposals}
%RW proposals are cheap but mix poorly in high dimensions \cite{mangoubi2018doeshamiltonianmontecarlo}.
The choice of MK impacts the quality of the posterior approximation. It has been shown that gradient-based MKs improve approximations in high-dimensional settings \cite{buchholz2020adaptive}. 

Hamiltonian Monte Carlo (HMC) \cite{neal2011mcmc} uses a gradient-based MK that simulates Hamiltonian dynamics for $S$ steps via the leapfrog scheme with step size $h$. Leapfrog can be considered a function which uses the state $\bm{\theta}_{k-1}$ and momentum $\mathbf{P}_{k-1}$ to transform the state i.e. ${\bm{\theta}_{k} = f_{\text{LF}}(\bm{\theta}_{k-1}, \mathbf{P}_{k-1})}$. At iteration $k-1$ of the SMC sampler, an initial auxiliary momentum variable $\mathbf{P}_{k-1}^0$ is drawn from $\mathcal{N}(0,\bm{M})$ with mass matrix $\bm{M}$ which is typically an identity matrix $\mathbf{I}$. At step $s$, leapfrog updates the state and momentum through
\begin{align}
    \mathbf{P}^{s + \frac{1}{2}}_{k-1} &= \mathbf{P}_{k-1}^s + \tfrac{h}{2}\nabla \log \pi(\bm{\theta}^s_{k-1}), \label{eq:mom_first} \\
    \bm{\theta}^{s+1}_{k-1} &= \bm{\theta}^s_{k-1} + h \mathbf{P}_{k-1}^{s + \frac{1}{2}}, \label{eq:pos_update} \\
    \mathbf{P}^{s+1}_{k-1} &= \mathbf{P}_{k-1}^{s + \frac{1}{2}} + \tfrac{h}{2}\nabla \log \pi(\bm{\theta}_{k-1}^{s+1}). \label{eq:mom_second}
\end{align}
After $S$ steps $\bm{\theta}^{S}_{k-1}=\bm{\theta}_{k}$ and $\mathbf{P}^{S}_{k-1}=\mathbf{P}_{k}$. We note that Langevin dynamics (LD) are another gradient-based MK, and can be viewed as an $S = 1$ leapfrog process~\cite{rosato2024enhancedsmc2leveraginggradient, murphy2025hessmc2sequentialmontecarlo}.

The L-kernel is a user-specified probability distribution, the optimal choice is generally intractable but approximations can be made \cite{green2022increasing}. An alternative is to use the MK as the L-kernel \cite{devlin2024NUTS}. When using MCMC kernels in SMC samplers, the accept-reject component ensures that the MK and L-kernel will cancel in the weight update \eqref{equ:weight_update}. However without accept-reject, we use a change of variables to account for the transformation of the state due to the sampled momentum. We can write the proposal for an MK which uses leapfrog as
\begin{equation}
    q_k(\bm{\theta}_k|\bm{\theta}_{k-1}) 
    = \mathcal{N}(\mathbf{P}_{k-1};0,\bm{M})
      \begin{vmatrix}
        \tfrac{df_{LF}(\bm{\theta}_{k-1},\mathbf{P}_{k-1})}{d\mathbf{P}_{k-1}}
      \end{vmatrix}^{-1}, \label{eq:q_prop}
\end{equation}
and the corresponding L-kernel is
\begin{align}
    L_k(\bm{\theta}_{k-1}|\bm{\theta}_k)
    &= \mathcal{N}(-\mathbf{P}_k;0,\bm{M})
      \begin{vmatrix}
        \tfrac{df_{LF}(\bm{\theta}_k,-\mathbf{P}^*)}{d\mathbf{P}^*}
      \end{vmatrix}^{-1}. \label{eq: l=q_deriv}
\end{align}
The determinants in \eqref{eq:q_prop} and \eqref{eq: l=q_deriv} cancel in \eqref{equ:weight_update} leaving
\begin{equation}
    \mathbf{w}^{(j)}_k = \mathbf{w}^{(j)}_{k-1} 
    \frac{\pi(\bm{\theta}^{(j)}_k)}{\pi(\bm{\theta}^{(j)}_{k-1})}
    \frac{\mathcal{N}(-\mathbf{P}^{(j)}_k;0,\bm{M})}{\mathcal{N}(\mathbf{P}_{k-1}^{(j)};0,\bm{M})},
    \qquad j=1,\dots,J. 
    \label{equ:weight_update_fp}
\end{equation}
Full kernel derivations for HMC and LD are in \cite{devlin2024NUTS} and \cite{murphy2025hessmc2sequentialmontecarlo}. 
\section{Batch Inference and Data Annealing}\label{sec:batch}

\begin{table*}[t]
    \tiny
    %\scriptsize
    %\footnotesize
    \centering
    \resizebox{1.0\textwidth}{!}{%
    \begin{tabular}{l l c c   c c  c c}
        \toprule
        \multirow{2}{*}{\textbf{Batching Scheme}} & \multirow{2}{*}{\textbf{Experiment}} 
        & \multicolumn{2}{c}{\textbf{Test Loss}} 
        & \multicolumn{2}{c}{\textbf{Test Accuracy (\%)}} 
        & \multicolumn{2}{c}{\textbf{Runtime (s)}} \\
        \cmidrule(lr){3-4} \cmidrule(lr){5-6} \cmidrule(lr){7-8}
         & & \textbf{LD} & \textbf{HMC} 
           & \textbf{LD} & \textbf{HMC} 
           & \textbf{LD} & \textbf{HMC} \\
        \midrule
        \multirow{2}{*}{Constant} 
            & MNIST & $-0.2726 \pm 0.003$ & $-0.081 \pm 0.004$ 
                    & $92.46 \pm 0.01$ & $97.55 \pm 0.11$ 
                    & $502.76 \pm 20.28$ & \bm{$453.33 \pm 16.74$} \\
            & FashionMNIST & $-0.489 \pm 0.010$ & $-0.345 \pm 0.011$ 
                            & $82.79 \pm 0.41$ & $88.09 \pm 0.26$ 
                            & \bm{$392.11 \pm 11.93$} & $443.58 \pm 12.70$ \\
        \midrule
        \multirow{2}{*}{FB} 
            & MNIST & $-0.248 \pm 0.002$ & \bm{$-0.065 \pm 0.003$} 
                    & $93.14 \pm 0.01$ & \bm{$98.00 \pm 0.15$} 
                    & $8985.56 \pm 24.08$ & $10658.68 \pm 15.28$ \\
            & FashionMNIST & $-0.468 \pm 0.011$ & $-0.293 \pm 0.006$ 
                            & $83.46 \pm 0.42$ & $89.60 \pm 0.20$ 
                            & $8749.95 \pm 25.22$ & $10864.92 \pm 33.06$ \\
        \midrule
        \multirow{2}{*}{CTR} 
            & MNIST & $-0.268 \pm 0.003$ & $-0.071 \pm 0.002$ 
                    & $92.61 \pm 0.10$ & $97.84 \pm 0.02$
                    & $1297.93 \pm 19.09$ & $1535.95 \pm 20.27$ \\
            & FashionMNIST & $-0.483 \pm 0.010$ & \bm{$-0.291 \pm 0.005$} 
                            & $82.99 \pm 0.55$ & \bm{$89.62 \pm 0.30$}
                            & $1265.94 \pm 13.71$ & $1531.53 \pm 12.13$ \\
        \midrule
        \multirow{2}{*}{Linear} 
            & MNIST & $-0.249 \pm 0.004$ & $-0.067 \pm 0.003$ 
                    & $93.14 \pm 0.12$ & $97.90 \pm 0.13$
                    & $7904.98 \pm 22.95$ & $9507.66 \pm 14.32$ \\
            & FashionMNIST & $-0.470 \pm 0.010$ & $-0.293 \pm 0.004$ 
                           & $83.47 \pm 0.41$ & \bm{$89.62 \pm 0.26$}
                            & $7575.59 \pm 24.90$ & $9578.16 \pm 21.47$ \\
        \midrule
        \multirow{2}{*}{Automated} 
            & MNIST & $-0.260 \pm 0.003$ & $-0.067 \pm 0.002$ 
                    & $92.79 \pm 0.13$ & $97.94 \pm 0.05$
                    & $4801.25 \pm 23.69$ & $5666.36 \pm 45.31$ \\
            & FashionMNIST & $-0.478 \pm 0.010$ & $-0.293 \pm 0.005$ 
                            & $83.16 \pm 0.52$ & $89.51 \pm 0.30$
                            & $4700.58 \pm 7.02$ & $5891.13 \pm 26.41$ \\
        \midrule
        \multirow{2}{*}{SDA} 
            & MNIST & $-0.279 \pm 0.005$ & $-0.074 \pm 0.006$ 
                    & $92.28 \pm 0.20$ & $97.70 \pm 0.17$
                    & $6872.41 \pm 114.60$ & $7683.07 \pm 83.69$ \\
            & FashionMNIST & $-0.493 \pm 0.005$ & $-0.298 \pm 0.005$ 
                            & $82.50 \pm 0.23$ & $89.35 \pm 0.10$
                            & $6594.19 \pm 32.74$ & $7897.29 \pm 98.47$ \\
        \bottomrule
    \end{tabular}}
    \caption{Test loss, accuracy, and runtime (mean $\pm$ std) across proposal methods and DA schemes.}
    \label{tab:image_loss_acc_runtime}
\end{table*}

Particle-based methods have been applied to the Bayesian inference of NN posteriors but are a computationally intensive approach~\cite{neal2012bayesian,DBLP:journals/corr/abs-2104-14421}. Methods such as stochastic gradient LD/HMC are faster as they use MB stochastic estimates of log-likelihoods (LLs) and gradients \cite{chen2014stochastic}. The stochastic gradient estimate is
\begin{align}
    \nabla \log p(\mathbf{y}_{1:N}\mid\bm{\theta})
    \;\approx\;
    \frac{N}{M_k}\,\sum_{i=1}^{M_k}\nabla \log p(\mathbf{y}_i\mid\bm{\theta}),
    \label{equ:approx-grad}
\end{align}
where $\mathbf{y}_{1:M_k}$ denotes the current batch of size $M_k$, drawn without replacement from $\{1,\ldots,N\}$. The gradient is then scaled by $\frac{N}{M_k}$ giving an approximation of the full batch gradient, a similar technique is used in VI \cite{graves2011practical}. A smaller $M_k$ yields faster, lower-memory estimates, while a larger $M_k$ improves estimate quality but increases memory usage. 

%especially when propagating many parameter samples in parallel (e.g.\ with \texttt{vmap} in JAX~\cite{jax2018github}). 

To balance estimate cost and quality, we adopt DA schemes. We begin with small batches to move samples quickly toward high probability mass regions and then increase the batch size to approach full batch accuracy in gradient estimates for better exploitation. At each annealing step we \emph{append one full MB of datapoints}, cumulatively growing the training subset. The LL at each iteration is calculated with previously included data and the newly added MB. We examine five naive schemes and one in which data is introduced according to an entropy-based criterion. The complexity of each scheme when using gradient-based proposals is $\mathcal{O}(\gamma(M_k) S)$ where $\gamma$ denotes the cost of a gradient evaluation for one datapoint.

\subsection{Annealing Schemes}
Let $C$ denote the initial batch size and $\kappa$ the MB increment (the number of datapoints added at each step). We write $M_k$ for the batch size used at iteration $k$, with $C \le M_k \le N$. The five naive schemes are as follows:

\begin{itemize}
    \item \textbf{Constant}: fixed MB.
    \begin{equation}
        M_k = C.
    \end{equation}
    \item \textbf{Full-batch (FB)}: always use all data.
    \begin{equation}
        M_k = N.
    \end{equation}
    \item \textbf{Constant-to-refine (CTR)}: simple two-stage schedule.
    \begin{equation}
        M_k =
        \begin{cases}
        C, &k < 0.9K,\\
        N, &\text{else.}
        \end{cases}
    \end{equation}
    \item \textbf{Linear}: add one MB per step until full.
    \begin{equation}
        M_k = \begin{cases}
            \kappa k +C, \quad &k<0.9K,\\
            N, \quad &\text{else.}
        \end{cases}.
    \end{equation}
    \item \textbf{Automated}: linearly increasing scheme that reaches FB at $0.9K$. In practice, $M_k$ is rounded to the nearest multiple of the MB size $\kappa$, since new datapoints are always appended in full MBs.
    \begin{equation}
        M_k =
        \begin{cases}
        C + \left\lfloor \dfrac{N - C}{0.9K}\ \right\rfloor k, &k < 0.9K,\\[6pt]
        N, &\text{else.}
        \end{cases}
    \end{equation}  
\end{itemize}

Linear and Automated reduce to the same schedule with certain parameter values. Figure~\ref{fig1:anneal_schemes} shows how $M_k$ evolves under each schedule. Constant and FB are baselines; CTR tests if a long fixed MB phase followed by a brief FB phase suffices for efficient sampling; and the two linear schemes explore gradual annealing via incremental MB additions.
    
\begin{figure*}[t]
    \centering
    \includegraphics[width=0.95\linewidth]{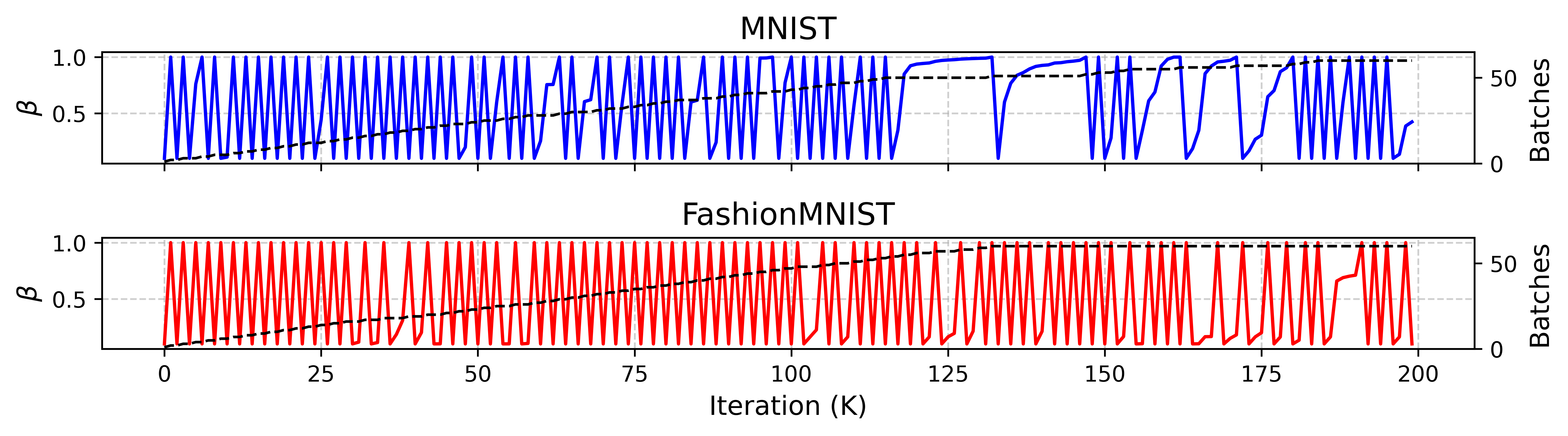} % path to your file
    \caption{Change in $\beta$ values and batch count over iterations for a single run on each dataset with the SDA scheme. Early on, $\beta_k$ often resets to 1, rapidly adding MBs, but the inclusion rate slows over time.}
    %\caption{Change in $\beta$ values and batch count over iterations for a single run on each dataset with the SDA scheme.}
    \label{fig:betas_comparison}
\end{figure*}

Our final batching scheme is smooth DA (SDA) \cite{green2015bayesian}. SDA introduces data into the LL of a target such that the Shannon entropy $\mathcal{S}$ varies at a constant rate. At iteration $k$, $M_{k-1}$ is the number of already annealed-in datapoints and the next MB to be annealed-in is $M_{k-1}+1:M_k$ . The negative LL contributions are
\begin{align}
    \Omega_k &= -\log p(\mathbf{y}_{1:M_{k-1}}\mid\bm{\theta}), \\
    \hat{\Omega}_k &= -\log p(\mathbf{y}_{M_{k-1}+1:M_k}\mid\bm{\theta}).
\end{align}
Starting from $M_0=C$, SDA targets
\begin{equation}
    \pi_{\beta_k}(\bm{\theta}\mid\mathbf{y}_{1:M_k})
    \;\propto\;
    p(\mathbf{y}_{1:M_k}\mid\bm{\theta})^{\beta_k} \, q_0(\bm{\theta}),\label{eq:sda_initial_target}
\end{equation}
and adapts $\beta_k$ to change $\mathcal{S}$ at a constant rate via
\begin{equation}
    \beta_{k+1} = 
    \beta_k - \frac{\Delta \mathcal{S}}{\mathrm{Var}(\hat{\Omega}_k)} ,
    \label{eq:beta_initial_data_k}
\end{equation}
subject to the constraint $\beta_k < \beta_{k+1} \le 1$ with $\beta_{k+1}$ clipped to $1$. We target \eqref{eq:sda_initial_target} with $M_{k}=M_{k-1}$ until $\beta_{k+1}=1$, when we \emph{append one MB} ($M_{k}=M_{k-1}+\kappa$) and now target
\begin{align}
    \pi_{\beta_k}(\bm{\theta}\mid\mathbf{y}_{1:M_{k}})\;\propto\; 
    p(\mathbf{y}_{1:M_{k-1}}&\mid\bm{\theta}) \times \nonumber \\ \,
    &p(\mathbf{y}_{M_{k-1}+1:M_{k}}\mid\bm{\theta})^{\beta_k} \, q_0(\bm{\theta}),
\end{align}
with update
\begin{equation}
    \beta_{k+1}
    \;=\;
    \beta_k \;-\; \frac{\Delta \mathcal{S}}{\mathrm{Var}(\hat{\Omega}_{k}) + \mathrm{Cov}(\Omega_{k},\hat{\Omega}_{k})}.
    \label{eq:beta_annealed_k}
\end{equation}
 If the unconstrained update in \eqref{eq:beta_initial_data_k} or \eqref{eq:beta_annealed_k} yields $\beta_{k+1}\le \beta_k$, we flip the sign of $\Delta \mathcal{S}$ (a hyperparameter we have set to 1).  After each subsequent iteration in which $\beta_k$ reaches $1$, we append an additional MB and reinitialise $\beta_{k+1}$ to a small value (here $0.1$). In an SMC implementation with normalised weights $\{\tilde{\mathbf{w}}^j\}_{j=1}^J$ and particles $\{\bm{\theta}^j\}_{j=1}^J$, the required moments at iteration $k$ are
\begin{align}
    \text{Var}(\hat{\Omega}_k) &= \mathbb{E}[\hat{\Omega}_k^2] - \mathbb{E}[\hat{\Omega}_k]^2 \\
    \mathbb{E}[\hat{\Omega}_k] &\approx \sum_{j=1}^J \tilde{\mathbf{w}}^j (-\log p(\mathbf{y}_{M_{k-1}+1:M_k}|\bm{\theta}^j)) \\
    \mathbb{E}[\hat{\Omega}_k^2] &\approx \sum_{j=1}^J \tilde{\mathbf{w}}^j (-\log p(\mathbf{y}_{M_{k-1}+1:M_k}|\bm{\theta}^j))^2 \\
    \text{Cov}(\Omega_k, \hat{\Omega}_k) &= \mathbb{E}[\Omega_k\hat{\Omega}_k] - \mathbb{E}[\Omega_k]\mathbb{E}[\hat{\Omega}_k]\\
    \mathbb{E}[\Omega_k] &\approx \sum_{j=1}^J \tilde{\mathbf{w}}^j(-\log p(\mathbf{y}_{1:M_{k-1}}|\bm{\theta}^j)) \\
    \mathbb{E}[\Omega_k, \hat{\Omega}_k] &\approx \sum_{j=1}^J \tilde{\mathbf{w}}^j (\log p(\mathbf{y}_{1:M_{k-1}}|\bm{\theta}^j)\times \nonumber \\
    &\qquad \quad \log p(\mathbf{y}_{M_{k-1}+1:M_k}|\bm{\theta}^j)) \label{eq:E{XY}}
\end{align}
%

%\subsection{Pseudo-Deep Ensembles via Weighted Voting}
%Deep ensembles  train multiple independent NNs with different initialisations or batches, then combine predictions (e.g., by majority vote) to improve accuracy and quantify uncertainty. 
%SMC Samplers resemble a weighted deep ensemble \cite{lakshminarayanan2017simplescalablepredictiveuncertainty, d2021repulsive}: at each iteration, NNs for each sample are trained on different mini-batches, and normalised weights yield a weighted mean estimate. The resampling step retains well-performing models, enabling iterative improvement akin to genetic algorithms.  

% \section{Experiments}\label{sec:experiments}
% All experiments were implemented in JAX~\cite{jax2018github} and run on an NVIDIA A100 GPU. Training was performed for 200 iterations over 5 random seeds, reporting mean and standard deviation of test accuracy and loss. The proposals used a step size of $h = 0.002$ with HMC using $S = 3$ leapfrog steps. The MB and initial batch sizes were $\kappa = C= 500$. The first task used the MNIST dataset \cite{lecun1998mnist}, consisting of 70,000 grayscale images of handwritten digits. We trained a LeNet-5 architecture \cite{726791} with 2 convolutional and 2 dense layers, using $\text{tanh}$ activations, giving $D=61,706$ parameters. The second used the FashionMNIST \cite{xiao2017fashionmnist} dataset, a benchmark of 70,000 grayscale images across 10 clothing classes. Here we used a larger CNN: 2 convolutional–batchnorm–maxpool blocks followed by 2 dense layers with $\text{tanh}$ activations, giving $D=96,658$ parameters. 

\section{Discussion and Conclusions}\label{sec:discussion}
All experiments were implemented in JAX and run on an NVIDIA A100 GPU. Training was performed for 200 iterations over 5 random seeds, reporting mean and standard deviation of test accuracy and loss. The proposals used a step size of $h = 0.002$ with HMC using $S = 3$ leapfrog steps. The MB and initial batch sizes were $\kappa = C= 500$. The first task used the MNIST dataset \cite{lecun1998mnist}, consisting of 70,000 grayscale images of handwritten digits. We trained a LeNet-5 architecture \cite{726791} which has $D=61,706$ parameters. The second used the FashionMNIST \cite{xiao2017fashionmnist} dataset, a benchmark of 70,000 grayscale images across 10 clothing classes. Here we used a larger CNN with $D=96,658$ parameters. 

%We introduced mini-batch gradient and likelihood estimation into SMC samplers for efficient BNN posterior inference. It was found that DA schemes provide speed-up with minimal performance loss compared to FB inference.

Table~\ref{tab:image_loss_acc_runtime} reports test loss, accuracy, and runtime for the image classification tasks. Overall, DA schemes show limited impact on convergence. FB performs slightly better at a much higher computational cost, while Constant is fastest, giving an $\approx20\times$ speed up but with a small drop off in accuracy. CTR emerges as a strong compromise: it matches FB performance during refinement while reducing runtime by a factor of 6.6. Note that the refinement period of CTR uses the full batch at the end, so the gradient estimates are not scaled approximations. Linear generally outperforms Automated and SDA. 

For FashionMNIST, more expensive schemes performed better overall, though CTR narrowed the gap once refinement began. This demonstrates that cheaper schemes can guide samples efficiently toward the target before switching to FB inference, reducing training time with little accuracy loss. Although FB marginally outperformed other batching schemes, its computational burden suggests that longer refinement within cheaper schemes may be a better trade-off for high-dimensional models.

Finally, across all schemes, LD consistently underperformed HMC. Even with a short leapfrog length ($L=3$), HMC offered superior overall results with only a slight increase in runtime. Future work could explore using adaptive trajectory length methods in proposals \cite{millard2025incorporatingcheescriterionsequential}. 

\vfill

\bibliographystyle{IEEEbib}
\bibliography{main}
%\bibliography{strings,refs}

%\input{7_appendix}

\end{document}